\documentclass[11pt,letterpaper]{article}

\usepackage[letterpaper,margin=1in,headheight=14pt,headsep=18pt]{geometry}
\usepackage[T1]{fontenc}
\usepackage[utf8]{inputenc}
\usepackage[american]{babel}
\usepackage{csquotes}
\usepackage{XCharter}
\usepackage[xcharter,vvarbb]{newtxmath}
\usepackage[scaled=0.9]{zi4}          
\usepackage{amsmath}
\usepackage{graphicx}
\usepackage[natbibapa]{apacite}
\usepackage{microtype}
\usepackage{xcolor}
\definecolor{accent}{gray}{0}
\definecolor{rulegray}{gray}{0}

\usepackage{titlesec}
\titleformat{\section}{\normalfont\Large\bfseries\color{accent}}{\thesection}{0.7em}{}
\titleformat{\subsection}{\normalfont\large\bfseries\color{accent}}{\thesubsection}{0.7em}{}
\titleformat{\subsubsection}[runin]{\normalfont\normalsize\bfseries}{}{0pt}{}[.]
\titlespacing*{\section}{0pt}{3.0ex plus 1ex minus .3ex}{1.2ex plus .2ex}
\titlespacing*{\subsection}{0pt}{2.4ex plus .8ex minus .3ex}{0.9ex plus .2ex}
\titlespacing*{\subsubsection}{0pt}{1.4ex plus .4ex minus .2ex}{0.6em}

\usepackage[font=small,labelfont={bf,color=accent},labelsep=period,
            justification=justified,singlelinecheck=false,skip=8pt]{caption}

\usepackage{fancyhdr}
\fancypagestyle{plain}{\fancyhf{}\fancyfoot[C]{\small\thepage}}

\usepackage{hyperref}
\hypersetup{colorlinks=true,linkcolor=accent,citecolor=accent,urlcolor=accent,
  pdftitle={Assessing the Effect of Cross-Domain Mapping on Creativity in Humans and Large Language Models},
  pdfauthor={Qiawen Ella Liu, Marina Dubova, Henry Conklin, Takumi Harada, Thomas L. Griffiths}}

\begin{document}
\thispagestyle{plain}

\begin{center}
{\LARGE\bfseries\color{accent} Assessing the Effect of Cross-Domain Mapping\\[3pt]
on Creativity in Humans and Large Language Models\par}
\vspace{1.4em}
{\normalsize Qiawen Ella Liu\textsuperscript{1},\enspace
Marina Dubova\textsuperscript{2,\textdagger},\enspace
Henry Conklin\textsuperscript{1,\textdagger},\enspace
Takumi Harada\textsuperscript{1,3},\enspace
Thomas L. Griffiths\textsuperscript{1}\par}
\vspace{0.7em}
{\small\itshape
\textsuperscript{1}Princeton University\par
\textsuperscript{2}University of California, Berkeley\par
\textsuperscript{3}Toyota Motor North America, Inc., Plano, Texas\par}
\vspace{0.5em}
{\small
\textsuperscript{\textdagger}Equal contribution\qquad
Correspondence: \href{mailto:ella.qiawen.liu@princeton.edu}{\texttt{ella.qiawen.liu@princeton.edu}}\par}
\end{center}

\vspace{1.2em}
\begin{center}
\begin{minipage}{0.9\textwidth}
{\color{accent}\rule{\linewidth}{0.8pt}}\par\vspace{6pt}
{\normalsize\bfseries\color{accent}Abstract}\par\vspace{4pt}
{\small\linespread{1.1}\selectfont
Creative ideas often arise by associating remote concepts. Can random associations reliably increase originality, and do they help humans and large language models (LLMs) in the same way? We asked human participants and seven LLMs to design products by drawing inspiration from a random source or addressing an unmet user need. Humans reliably benefited from cross-domain mappings, while LLMs generated more original ideas than humans but showed no overall benefit from the intervention, though this changed with semantic distance. More distant source-target pairings produced more original ideas in both humans and LLMs. Humans benefited at nearly any distance, while only the highest-rated LLMs benefited when the source was sufficiently remote. Humans and LLMs also used sources differently: humans tended to transfer surface features, while LLMs transferred structural and functional properties. These findings reveal the generative role of remote associations and systematic differences in how humans and AI respond to the same creativity intervention.\par}
\vspace{8pt}
{\small\textbf{\color{accent}Keywords:} creativity, analogy, large language models, product design\par}
\vspace{4pt}{\color{accent}\rule{\linewidth}{0.4pt}}
\end{minipage}
\end{center}
\vspace{1em}

\section{Introduction}
Coming back from a hike in 1941, Swiss engineer George de Mestral noticed burrs stuck stubbornly to his dog's fur. Upon close examination of these burrs with a microscope, he found numerous tiny hooks catching on loops---a structure that inspired his invention of Velcro, the fastener now securing everything from toddlers' shoes to spacesuits \citep{Stephens2007Velcro}. Similar cross-domain transfers appear throughout the history of invention: Early aircraft were inspired by the structure of bird wings \citep{Lilienthal1889}, and neonatal incubators were inspired by temperature-regulation techniques from poultry brooders \citep{budin1907nursling}. In each case, invention comes not from nothing but from mapping existing concepts across domains.

Drawing inspiration across domains may appear more like a game of chance than a systematic method for inducing creativity, given it depends on the serendipitous combination of a specific hike with a specific dog at a specific place.
In the general case, what makes a cross-domain mapping successful, and can it be engineered as a reliable intervention to enhance creativity?
Existing research suggests humans might benefit from cross-domain inspirations \citep{chan2011benefits,dahl2002influence,wilson2010effects}. The reason is that humans are notoriously prone to fixation: once a problem is framed in a familiar way, people tend to reuse known solutions within a single domain rather than scanning for distant alternatives across domains \citep{basalla1988evolution, jansson1991design,german2005functional}. Even when people try to overcome fixation by actively considering alternatives, they can still be constrained by knowledge fragmentation \citep{simon1955behavioral}, characterized by the classic `unknown unknown' problem \citep{firestein2012ignorance}.

Large language models (LLMs) provide a valuable comparison for asking whether cross-domain mapping facilitates idea generation across different kinds of intelligent systems, or whether it is especially useful because it helps overcome constraints specific to human cognition. LLMs are trained on massive, cross-disciplinary corpora that might allow them to generate remote associations without falling prey to the same cognitive bottlenecks. While a central theme of human creativity research is developing scalable methods that help innovators overcome both fixation and unknown unknowns, it remains an open question whether interventions designed for humans will prove equally effective for systems with fundamentally different capacities for storing, retrieving, and recombining conceptual knowledge.

Here, we test cross-domain mapping as a practical intervention for idea generation by both humans and LLMs. We prompt both to generate novel product ideas either by explicitly drawing inspiration from a distant source domain or by proposing incremental improvements based on existing user needs. Human raters then assessed the ideas along four dimensions commonly associated with creative products: originality (our focal dimension), feasibility, utility, and overall investment-worthiness. Evaluating these dimensions separately, rather than collapsing them into a single creativity score, lets us examine how cross-domain mapping affects different components of creative evaluation and, as we will show, the same intervention can move these components in opposite directions. We ask whether cross-domain mapping reliably boosts originality of ideas, whether its effects differ between humans and LLMs, and what kinds of these mappings are generative in both systems.

We show that cross-domain mapping significantly boosts the originality of ideas for human participants but not the tested LLMs, and that LLMs produce more original ideas than humans on average. However, this does not mean cross-domain mapping is useless for LLMs across the board. We show that the benefit of cross-domain mapping grows as the distance between source and target gets larger, and while human participants in our study benefit from almost any source, the highest-rated LLMs benefit only from the more distant half of the pairings. Finally, we reveal a qualitative difference in the information the two systems map from an inspiration source: humans tend to transfer surface features of a semantic concept, while LLMs tend to transfer a concept's underlying structures and functions. Overall, our results highlight the generative role of remote association and systematic differences in how humans and LLMs engage in creative generation.

\subsection{Creativity as the Ability to Bridge Remote Associations}

Creativity researchers have long recognized that creative ideas often arise by linking concepts that are \textit{weakly associated}, traversing longer distances in semantic space or recombining elements drawn from distinct semantic domains \citep{mednick1962associative,koestler1964act,kenett2018going,youn2015invention,green2016creativity}. Evidence for this relationship emerges both across ideas and across individuals. Across a range of divergent-thinking and associative tasks, responses that are semantically more distant from the object or cue that elicited them tend to be perceived as more original \citep{beaty2022semantic,heinen2018semantic,beaty2021automating,dumas2021measuring}. Highly creative individuals also tend to navigate semantic space more expansively, retrieving more distantly related concepts \citep{beaty2021forward,gray2019forward,kenett2014investigating,benedek2013revisiting}, shifting flexibly across semantic subcategories \citep{zhang2023retrieval}, and producing larger associative leaps both with and without explicit goals (\citealp[see][for a review]{beatyKenett2023associative}).

Research on human creativity suggests that the ability to make remote associations may depend both on the knowledge represented in semantic memory and on how that knowledge is searched \citep{beaty2014roles, beaty2021forward, beaty2023semantic}. Associative theories emphasize the structure of semantic memory, including what associations are available and how they are organized \citep{mednick1962associative, kenett2014investigating, kenett2019semantic, benedek2023role, kenett2018flexibility}. \citet{mednick1962associative} proposed that creativity benefits from having more associations and a flatter associative hierarchy, in which remote associates remain accessible rather than being dominated by a few conventional responses. Later network models extend this idea, linking creativity to network
properties such as shorter paths between concepts, lower modularity, and greater small-world organization \citep{kenett2014investigating, benedek2017semantic, kenett2019semantic, he2020relation, kenett2018flexibility}. Executive theories, on the other hand, emphasize how knowledge is searched. More creative individuals may be better able to move beyond dominant responses and fixate less on obvious associations \citep{benedek2013revisiting, benedek2014intelligence, beaty2014roles}. Consistent with this view, executive abilities such as inhibition and updating predict creative performance \citep{benedek2014intelligence}, and different retrieval dynamics can produce different search patterns over the same semantic structure \citep{hills2025entropy,hills2012optimal}.

These accounts of creativity suggest remote associations may fail to arise if the concepts are weakly connected or difficult to access within semantic memory, or search may terminate before reaching them. One way to overcome these constraints is to supply a remote association rather than requiring the creator to discover it through free search. Previous studies have shown that externally supplied remote cues or analogies can improve creative outcomes in some contexts \citep{linsey2010study,zhan2023effects,wise2024sparking,ritter2017enhancement}, but have typically tested only a limited set of cues or source--target combinations. We therefore use cross-domain mapping as a systematic intervention to explore the much larger combinatorial space of possible inspirations by \textit{randomly} sampling source--target pairs across domains. This use of random exploration also echoes recent work showing that less theory-guided sampling can reveal regions of a search space that more directed strategies may overlook \citep{dubova2026against}. In doing so, we hope to intentionally recreate the kind of serendipity that has been shown to be so important in the history of discovery \citep{souriau1881theorie,james1880great}.

\subsection{Analogy as the Mechanism for Transferring Relational Structure Across Domains}

Once a remote source of inspiration is supplied, the creator has to determine \textit{how} that source can be used in ideation. Analogy provides one powerful strategy for structuring the creative process by constraining the ways in which the source can be mapped onto the target \citep{green2024process}. Analogy can support systematic inference by aligning relational structure between a source and a target---importing a coherent system of relations that makes previously overlooked aspects of the target problem visible \citep{gentner1983structure,holyoak1995mental,holyoak2025human,holyoak2024analogy}. In particular, aligning with and projecting from the source structure can highlight implicit limitations in the target (constraints) as well as suggest new possibilities for intervention (affordances). For example, by mapping the \emph{circulatory system} onto \emph{urban traffic flow} \citep{kaplan2007membrane,chen2012model}, a planner might notice the need for one-way ``valves'' to prevent backflow, or see an opportunity to build lots of small neighborhood routes that deliver traffic deep into side streets, the way capillaries carry blood into tissue. 

Much of the work on analogy in creative design begins with a predefined problem and asks whether a source can help solve it \citep{yang2022creative,goel1997design,hey2008analogies,ozkan2013cognitive,goucher2019neuroimaging,goucher2019crowdsourcing}. For example, participants might be asked how to make an athletic shoe less slippery and draw on previously learned information about an octopus's suction cups to generate a solution \citep{yang2016prototypes}. To identify useful analogies on a larger scale, some work has used crowds and computational methods to search large repositories for existing designs that might help solve a target problem \citep{yu2014searching,hope2017accelerating,gilon2018analogy,kittur2019scaling}. Across these approaches, the mapping is typically guided by some prior specification of the problem to solve or the kind of solution to look for.

However, analogy can also be generative in the problem space itself. Mapping a source onto a target can reveal constraints and affordances that were hidden, expanding not only the range of possible solutions but also what can be conceived as a problem at all. This generative role of analogy has long been recognized in scientific discovery, where analogical models can introduce new ways of representing a phenomenon and support new inferences \citep{gentner1993shift,nersessian2008creating}, and scientists have been observed using structural analogies during hypothesis generation to project relations from one biological system onto another and identify new relations to investigate \citep{dunbar2001invivo}. Our intervention highlights this generative role in product design by randomly assigning sources rather than selecting them to address a predefined problem.

Not all mappings are equally generative. The analogy literature distinguishes \emph{surface} mappings, which transfer perceptual attributes (e.g., the sun and an atom's nucleus are both round and bright), from \emph{structural} mappings, which transfer systems of interconnected relations (e.g., a central body attracts smaller orbiting bodies) \citep{gentner1997structure,holyoak1987surface}. Developmental work documents a \emph{relational shift}: children initially map shared surface attributes, and only later develop a preference for systems of mutually constraining relations over isolated features \citep{gentner1988metaphor,rattermann1998more,gentner1986systematicity,clement1991systematicity,markman_structural_1993}. Yet successful structural mapping remains hard even for adults, who are readily distracted by surface similarities. For example, after reading a story about a hawk who stops a hunter's attacks by giving him some of her feathers, people are more likely to be reminded of it by another story with surface-level matches (e.g., stories about an eagle and a sportsman that share similar characters) rather than structural matches (e.g., a story about a small country that stops a neighbor's missile attacks by offering to sell it computers), even though they rate the structural matches as more sound \citep{gick1980analogical,keane1987retrieving,gentner1993roles,holyoak1987surface,ross1987this,gick1983schema}.

Overcoming this surface bias appears to require substantial domain expertise: For example, asked to complete the analogy ``a floating balloon is like \_\_ because \_\_,'' novices typically retrieve things that share a surface feature with the balloon, e.g., a kite, a leaf, other objects ``in the air.'' Expert geoscientists, by contrast, overwhelmingly retrieved analogies grounded in the mechanism that makes a balloon rise, like other systems in which something less dense rises through a denser medium, such as warm water rising in a cold sea \citep{goldwater2021analogy}. In addition to domain expertise, the capacity for identifying the more abstract kind of relational similarities is supported by the distributional properties of natural language \citep{gentner2010mutual,loewenstein2005relational,christie2014language}: shared labels and relational terms recur across contexts, implicitly linking otherwise distinct domains and making higher-order structure available for comparison and transfer.

\subsection{LLMs Can Generate Original Cross-Domain Associations at Scale}

Although whether LLMs are creative in the same sense as humans remains debated, particularly given their lack of the intentions and self-expression often associated with human creativity \citep[e.g.,][]{runco2023artificial,runco2025updating}, there is now substantial evidence that they can generate ideas that human judges consider highly creative. Across creative writing, design, and hypothesis generation tasks, LLMs reliably produce novel ideas and, in many studies, match or outperform human participants \citep{si2024can,cropley2023artificial,bellemare2026divergent,koivisto2023best}. Recent evidence further shows that LLMs can match or outperform humans in analogical reasoning \citep{ding2023fluid,webb2023emergent,liu2026large} and flexibly recombine existing knowledge to generate novel solutions \citep{mehrotra2024enhancing,gu2024llms,mizrahi2026cooking}. Even though the most creative humans can still outperform current models \citep{koivisto2023best,wang2026large,bellemare2026divergent} and LLM outputs tend to be more homogeneous than human outputs \citep{moon2025homogenizing,wenger2026homogeneously}, evidence nevertheless suggests contemporary LLMs can generate outputs perceived as highly creative.

This raises the question of whether a cross-domain mapping intervention will have the same effect on LLMs as on humans, and there are reasons to expect either outcome. The scaffolds that support analogical creativity in humans (e.g., extensive domain knowledge and the abstract relations encoded in language) are central to what LLMs learn during pre-training. LLMs may therefore be especially capable of using a supplied source to construct productive mappings between otherwise distant domains, in which case they should benefit from the intervention like humans do. At the same time, training on massive, cross-disciplinary corpora may already allow LLMs to reach distant concepts without being prompted to, so that many supplied sources fall within what a model would access on its own. In that case, the benefit of supplying a source might be diminished and emerge only for sources more remote than those the models can readily reach. To explore these possibilities, we directly compare how humans and LLMs respond to the same intervention.

\subsection{The Current Study}

We elicited ideas for novel product designs from both humans and LLMs under two matched prompting conditions. In the \emph{cross-domain mapping} condition, participants (human or LLM) were given a target product (e.g., \emph{backpack}) and a randomly assigned inspiration source (e.g., \emph{cactus}), and instructed to translate a property of the source into a novel feature of the target. In the \emph{user-need} condition, the same targets were presented without an inspiration source; participants instead identified an unmet user need and proposed a feature addressing it. Ideas from both sources were then rated by a separate pool of human judges on originality, feasibility, utility, and investment-worthiness. We tested seven LLMs spanning major model families (\texttt{o3}, \texttt{Claude-Sonnet-4}, \texttt{gpt-4o}, \texttt{gemini-2.5-pro}, \texttt{deepseek-v3}, \texttt{Llama-3.3-70b}, \texttt{Qwen2.5-72B}). Full materials, procedures, and analytic details are reported in the Methods section.

\section{Methods}

\subsection{Eliciting Novel Ideas From Humans and LLMs}

\subsubsection{Participants}
We recruited online English-speaking adults from Prolific, retaining a final sample of $N = 140$ participants after screening\footnote{We manually reviewed all responses for AI-like phrasing and atypical response consistency, and cross-referencing with AI-detection software.} (56 male, 83 female, 1 non-binary). Participants' mean age was 43.24 years ($SD = 13$, range = 21--77).

\subsubsection{Target Products}
We selected ten everyday consumer products spanning diverse categories: \textit{smartphone, TV} (electronics), \textit{sofa, desk} (furniture), \textit{car} (vehicle), \textit{refrigerator} (appliances), \textit{backpack} (accessories), \textit{yoga mat} (sports equipment), \textit{chef's knife} (kitchen tools), and \textit{sneakers} (clothing).

\subsubsection{Inspiration Sources}
We also selected 26 inspiration sources across various domains: animals (octopus, chameleon), plants (bamboo, cactus), natural systems (coral reef, beehive), technological systems (hydroelectric dam, GPS navigation), physical phenomena (water cycle, tornado formation), abstract concepts (storytelling), social/organizational systems (symphony orchestra, hospital, government), and food (pickle, sandwich). Specifically, the ten target products could also serve as inspiration sources, provided they were not paired with themselves.

\subsubsection{Design and Procedure}
Participants were randomly assigned to one of two conditions and completed 10 design challenges, generating one novel feature for each product.
In the \textit{Cross-Domain Mapping} condition, each target product (\textit{e.g., mattress}) was paired with a randomly sampled inspiration source (\textit{e.g., toothbrush}). Participants first described a property of the source (e.g., \textit{toothbrush bristles reach tiny spaces}) and then proposed a product feature that translated that property to the target (\textit{a mattress layer of ultrafine bristles that gently brushes away dead skin while you sleep}). From the 10 targets and 26 inspiration sources, we sampled 100 target--source pairs (see Supplemental Material, Figures S5--S6, for the full set of combinations and their mean originality ratings for humans and LLMs). Each of the 100 target--source pairs was completed by 7 participants, yielding $700$ human-generated cross-domain ideas.
In the \textit{User-Need} condition, participants saw the same target products (e.g., mattress) without a source; instead, they identified an unmet user need (\textit{e.g., partners disturb each other by moving during sleep}) and proposed an innovative solution (\textit{e.g., separate mattress zones that absorb motion and automatically adjust firmness}). Each product was received by 70 participants, yielding $700$ human-generated user-need ideas.

\subsubsection{Rephrasing and Format Normalization}
To control for superficial differences in style, framing, and length between ideas generated by humans and LLMs, we normalize each idea by asking the same LLM (\texttt{o3}) to rephrase the original content into a short (1--2 sentence), reader-friendly product description. Rephrased outputs omit both the cross-domain analogy framing and explicit user-need framing, focusing only on what the product is and does.\footnote{For example, the idea ``A cactus has spines that collect moisture from the air and tissue that stores water. Like a cactus, this backpack uses micro-spines to condense moisture as you walk and channels it into an internal reservoir, providing water and cooling during long trips.'' is rephrased as ``This backpack collects moisture from the air as you move and channels it into an internal reservoir. The collected water can be used for hydration and provides gentle cooling during extended use.''} A pilot study that did not apply this rephrasing step reproduces the model-level ordering and the distance--originality relationship reported in the main analyses (see Supplemental Material, Section~S6).

\subsection{Evaluating Ideas}

\subsubsection{Participants}
We recruited $N = 1002$ online English-speaking adults from Prolific (459 male, 528 female, 11 non-binary, 4 preferred not to say). Participants' mean age was 41.9 years ($SD = 12.9$, range = 20--80). This study was preregistered at Open Science Framework (\href{https://osf.io/fqavu/overview?view_only=65eca78986674d4aa77c38444af31f3c}{link}).

\subsubsection{Material}
We split all rephrased human- and LLM-generated ideas ($n = 2800$) into 100 lists, each containing approximately 30 ideas sampled evenly across generation source (human vs.\ LLM), condition (cross-domain vs.\ user-need), and target product. Each idea was rated by approximately 10 independent participants. Each list also included four catch trials to assess attention and understanding of the rating scales. 108 (about 10\%) participants who failed 50\% or more of these checks were excluded.

\subsubsection{Procedure}
Participants were instructed to imagine themselves as judges evaluating startup pitches on a show like \textit{Shark Tank}. They rated each idea on four dimensions using 5-point Likert scales with clearly labeled endpoints, with the order of these ratings randomized between participants: \textit{Originality}: the extent to which the idea was novel or unexpected. \textit{Feasibility}: the degree to which the idea could plausibly be implemented with current technology. \textit{Utility}: the extent to which the idea would provide meaningful real-world benefits to users. \textit{Investment-Worthiness}: an overall judgment of whether the idea was strong enough to merit investment. Participants were instructed to rely on their intuitive judgments and to evaluate each idea independently. Ideas were presented in randomized order within each list.

\section{Results}

\subsection{Originality, Feasibility, Utility, and Investment-Worthiness}

\begin{figure}[t!]
\centering
\includegraphics[width=\textwidth]{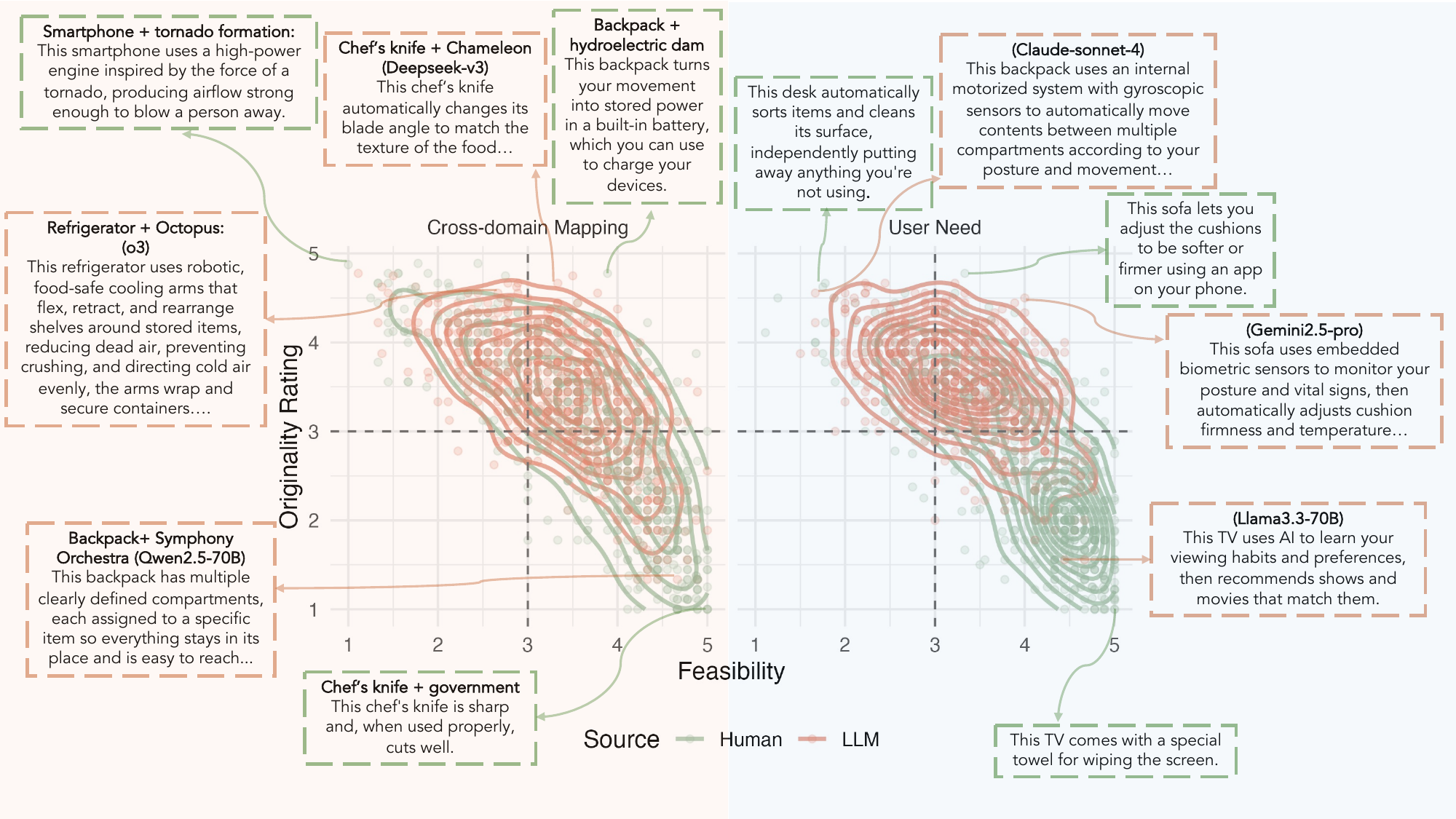}
\caption{Cross-domain mapping increases originality for humans but yields weaker benefits for LLMs. Ideas (points) are presented along originality (vertical axis) and feasibility (horizontal axis), with density contours showing distributions for human (green) and LLM (orange) outputs. Cross-domain prompting (left) produces more highly original human ideas relative to the user-need prompt (right).}
\label{fig:tradeoff}
\end{figure}

Across ideas, originality shows a strong trade-off with feasibility ($r = -0.74$, $p < .001$, as shown in Figure~\ref{fig:tradeoff}), and is weakly negatively correlated with utility ($r = -0.08$, $p < .001$). More original ideas are only modestly more likely to be judged as worth investing in ($r = 0.23$, $p < .001$). Investment-worthiness is driven primarily by perceived utility ($r = 0.71$, $p < .001$), far more than by originality or feasibility ($r = .11$, $p < .001$). The same qualitative relationships hold when considering human-generated ideas or LLM-generated ideas alone (see Supplemental Material, Figure S1). To assess the generalizability of the idea-level ratings across raters, we repeatedly divided the raters for each idea into two equal groups (2,000 random splits) and compared the average ratings from the two groups across ideas. Adjusting for the number of raters in each half \citep{spearman1910correlation,brown1910some,parsons2019psychological}, split-half reliability was $r = .81$ for originality, $.80$ for feasibility, $.73$ for utility, and was lower for investment-worthiness ($.56$). Overall, evaluations were generally consistent across different subsets of raters.

\subsection{Humans and LLMs Differ Across All Four Dimensions}

We assessed each of the four rating dimensions using a linear mixed-effects model predicting originality ratings from each source of ideas, with random intercepts for raters (since each rater rated multiple ideas) and ideas (since each idea was rated by multiple raters). Across both conditions, LLM-generated ideas were rated as more original, more useful, and more worthy of investment than humans, while human-generated ideas were rated as more feasible. We unpack the results on originality in the subsequent section and report full distributions and statistics for the other three dimensions in the Supplemental Material (Figures S2--S4).

\begin{figure}[t!]
\centering
\includegraphics[width=0.92\textwidth]{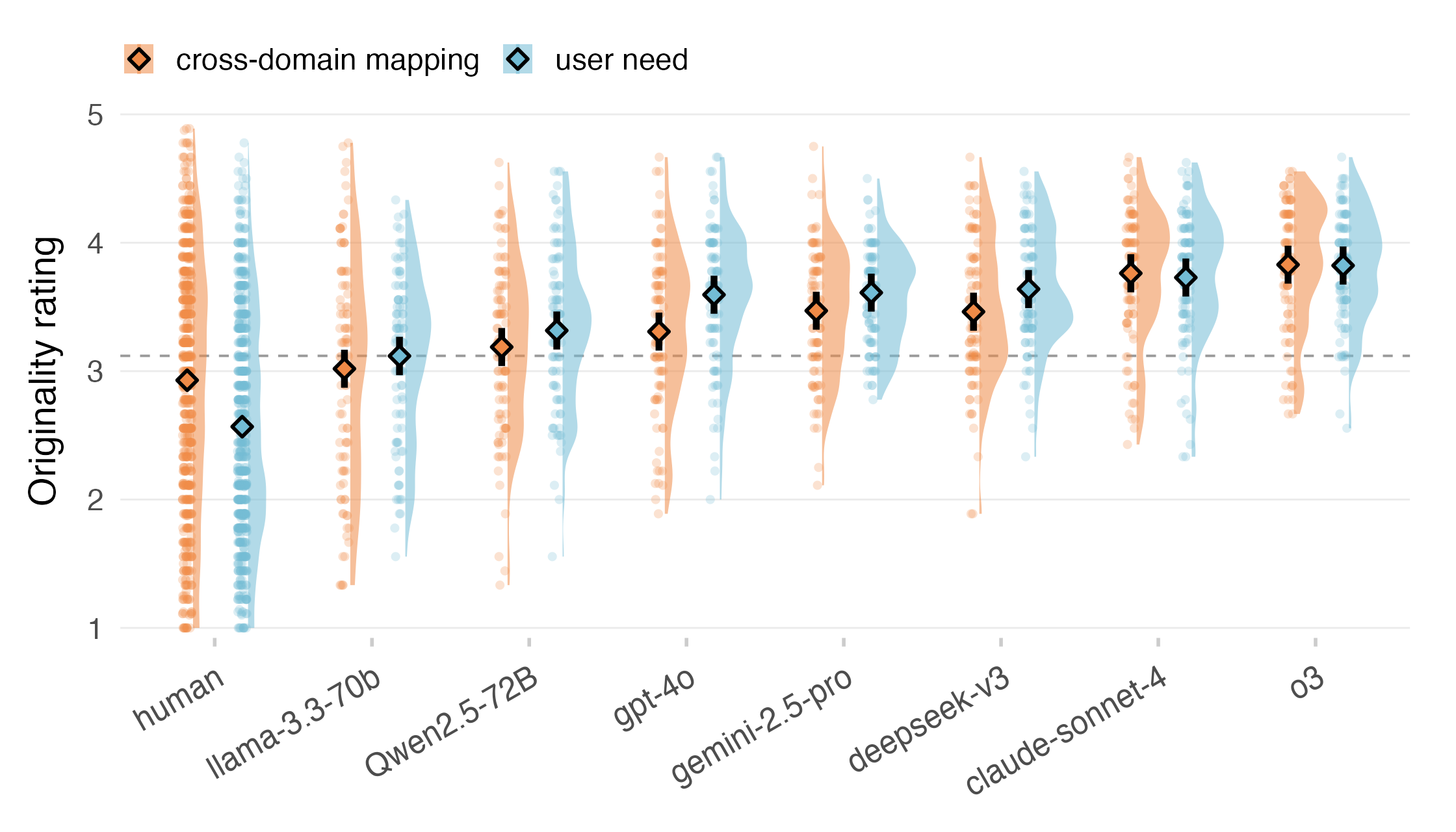}
\caption{Originality ratings by model and condition. Distributions show originality ratings for ideas generated by different models under each prompting conditions. Diamonds indicate estimated marginal means, with error bars showing 95\% confidence intervals.}
\label{fig:main_result}
\end{figure}

\subsection{Meaningful Differences in Originality Between Humans and LLMs, and Among LLMs}

Across both conditions, LLM-generated ideas were rated as more original than their human counterparts (see Figure \ref{fig:main_result}). Although cross-domain prompting increased originality for humans relative to the user-need condition, human ideas in the cross-domain mapping condition were still rated as less original than LLM-generated ideas overall ($M = 2.93$, 95\% CI $[2.87, 2.99]$). Most LLMs were rated as statistically significantly more original than humans after correction for multiple comparisons (Holm-adjusted $p < .01$), with the exception of \texttt{Llama-3.3-70b}, which did not reliably outperform humans ($M = 3.02$, 95\% CI $[2.87, 3.17]$). Among LLMs, originality ratings differed significantly: \texttt{o3} and \texttt{Claude-Sonnet-4} achieved the highest ratings ($M = 3.83$, 95\% CI $[3.68, 3.98]$ and $M = 3.76$, 95\% CI $[3.61, 3.91]$, respectively), significantly exceeding humans and other LLMs (all Holm-adjusted $p < .001$). \texttt{o3} and \texttt{Claude-Sonnet-4} did not differ reliably from each other, together forming a top-tier group that we revisit separately below, where we examine how the semantic distance between target and source impact originality.

A similar pattern held in the user-need condition. Human ideas again showed the lowest originality ratings ($M = 2.57$, 95\% CI $[2.50, 2.63]$), with all LLMs rated as more original than humans (all Holm-adjusted $p < .001$). However, unlike in the cross-domain mapping condition, differences among LLMs were smaller. Descriptively, \texttt{o3} and \texttt{Claude Sonnet 4} continued to receive the highest originality ratings; however, its advantage was statistically significant only relative to \texttt{Llama-3.3-70b} and \texttt{Qwen2.5-72B} (all Holm-adjusted $p < .001$), and not relative to \texttt{gpt-4o}, \texttt{gemini-2.5-pro}, or \texttt{deepseek-v3} after correction. The narrower spread among LLMs suggests that frontier models perform similarly when generating ideas from user needs, with clear differences emerging only under the more challenging cross-domain mapping task.

\subsection{Cross-Domain Mappings Increase Originality for Humans, but Not for LLMs}

We examined whether cross-domain prompting increases originality by fitting a linear mixed-effects model predicting originality ratings from idea source (human vs.\ LLM), condition (cross-domain mapping vs.\ user need), and their interaction, with random intercepts for participants and ideas. Aggregating across ideas, cross-domain mapping led to significantly higher originality ratings overall ($b = 0.36$, $SE = 0.04$, $t = 9.05$, $p < .001$). However, the augmentation effect of cross-domain mapping was primarily driven by human responses, as we also found a significant interaction between condition and idea source ($b = 0.48$, $SE = 0.06$, $t = 8.42$, $p < .001$). Specifically, for humans, cross-domain prompts reliably increased originality relative to the user-need condition, whereas LLM originality showed little evidence of a comparable benefit (see also Figure \ref{fig:tradeoff} and Figure \ref{fig:main_result}). However, this null effect for LLMs on average masks important heterogeneity across combinations and models. As we show in the next section, cross-domain mapping \textit{does} boost originality for the most capable LLMs when the inspiration source is sufficiently semantically distant from the target product.

The augmentation effect of cross-domain mappings for humans is even more pronounced in the upper tail of the originality distribution. When we focus on the top 10\% original ideas of each idea pool (human/LLM) ($n=96$), human ideas are slightly \emph{more original} than LLM ideas ($b=-0.07$, $SE=0.03$, $t=-2.19$, $p=.031$). For example, when asked to map \emph{tornado formation} onto a novel feature of smartphones, one human envisioned a phone that generates airflow strong enough to blow a person away (Figure~\ref{fig:tradeoff}, upper left corner); another imagined a smartphone which ``contains powerful internal fans that, when you press a button after misplacing it, spin to create a cyclone that guides the phone back to you.'' Inspired by \emph{GPS navigation}, someone proposed a shoe that ``shows your GPS route in bright colors, uses a light-up arrow on the top to point the way, and displays street names, road signs, and lights around the shoe.'' Across all ideas, these human ideas were rated as the most original (4.88--4.89 out of 5).

\begin{figure}[t!]
\centering
\includegraphics[width=\textwidth]{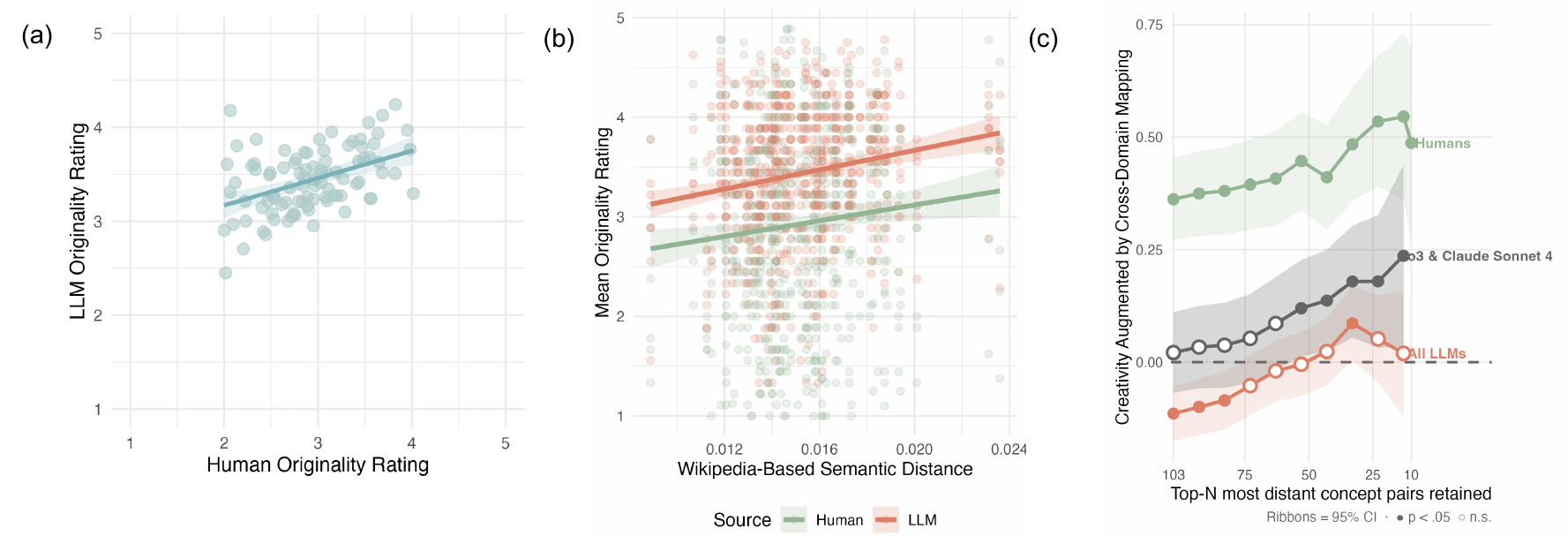}
\caption{Originality varies by cross-domain combinations and semantic distance: (a) originality ratings for human vs LLM ideas given the same target-source combinations are significantly correlated. (b) Mean originality ratings plotted against Wikipedia-based semantic distance between source and target concepts by source (human vs LLM). (c) Estimated benefit of cross-domain mapping over the user-need baseline as progressively closer target--source pairs are excluded; the leftmost point includes all pairs, and moving rightward restricts the analysis to increasingly distant subsets. Ribbons show 95\% CIs; filled points indicate $p < .05$, open points n.s.}
\label{fig:distance}
\end{figure}

\subsection{What Combinations Tend to Yield More Original Ideas}

Not all target--inspiration combinations were equally effective at inducing original ideas. Some pairings consistently yielded higher originality ratings across both humans and LLMs (see Figures S5 and S6). For instance, combining \textit{octopus} with \textit{car} produced original ideas from both humans and LLMs (e.g., human: ``This car moves on eight tentacle-like appendages instead of wheels, letting it drive, climb, sink, and float''; \texttt{o3}: ``like octopus, this car deploys a flexible skin embedded with micro-vacuum pads that [...] create temporary adhesion, allowing self-parking on steep walls''). In contrast, pairing \textit{sneaker} with \textit{sofa} resulted in mundane ideas that transferred cushioning features (e.g., human: ``This sneaker has extremely cushioned, plush soles designed specifically for comfort''; \texttt{Qwen2.5-72B}: ``this sneaker uses a multi-layered cushioning system ...''). The correlation between average originality ratings for the same combinations produced by humans versus LLMs was significantly positive ($r = 0.44$, $p < .001$, see also Figure \ref{fig:distance}a), suggesting that certain combinations are inherently more conducive to original thinking.

Prior work on divergent thinking suggest that responses farther from their eliciting cue in semantic space tend to be judged as more original \citep{beaty2022semantic,heinen2018semantic,beaty2021automating,dumas2021measuring,kenett2019what}. We therefore asked whether cross-domain mappings are more original when the supplied target and source are more semantically distant. To test this, we quantified the semantic distance between each pair using general knowledge associated with the corresponding concepts in Wikipedia. For each concept, we retrieved up to 50 sentences from its corresponding Wikipedia article (e.g., the ``Octopus'' article for \textit{octopus}) $X$
and embedded these sentences using a pretrained sentence-transformer model $\theta$ (\texttt{all-mpnet-base-v2}). We use the density estimation technique from \citet{conklin2025information}, which takes the embeddings $Z=\theta(X)$ and computes a soft quantization $\hat{Z}$ for each token by computing the cosine similarity between each embedding and $n$ uniformly distributed points on the surface of the unit sphere $S$ and normalising. The result is a categorical distribution $\hat{Z}$ over the discrete points in $S$ that describes the model's representation space:

\begin{equation}
       \underset{1 \times n}{P(\hat{Z})} \propto \sum \text{softmax}\biggl(\underset{bs \times h}{\frac{Z}{|Z|}} \cdot \underset{h \times n}{\frac{S}{|S|}}\biggr)
\end{equation}

By conditioning these estimates $P(\hat{Z}| X=x)$, we get a distribution for each concept $x$ describing how it is represented in the model's embedding space. Semantic distance between a target $i$ and its inspiration $j$ was computed as the Jensen--Shannon divergence between their distributions $D_{KL}\bigl(P(\hat{Z} \mid \text{concept}_i) \,\|\, P(\hat{Z} \mid \text{concept}_j)\bigr)$ at each layer, then aggregated. Across ideas, combinations that paired more semantically distant concepts tended to receive higher originality ratings (human $\beta=.1$, $t = 2.721$, $p < .01$; LLMs $\beta = .18$, $t = 4.81$, $p < .001$) (see Figure~\ref{fig:distance}b).

Given that semantic distance predicts originality, a natural question is whether the null effect of cross-domain mapping on LLM still holds once we account for the distance of the pairing. We progressively excluded low-distance combinations and compared originality in the cross-domain mapping condition against the user-need baseline for three groups: humans, all LLMs, and the two highest-performing LLMs (\texttt{o3} and \texttt{Claude-Sonnet-4}) (see Figure~\ref{fig:distance}c). Across all three groups, progressively restricting the analysis to more distant pairings gradually increased the benefit of cross-domain mapping. For humans, the effect was already large and significant when all pairings were included, and grew larger still as closer pairings were excluded. Pooled across LLMs, cross-domain mapping initially \textit{underperformed} the user-need condition, but the gap narrowed as closer pairings were excluded and the difference between two conditions was no longer significant once the closest 20\% of pairs were excluded. For \texttt{o3} and \texttt{Claude-Sonnet-4}, when all pairings were included, there was no significant effect for cross-domain mapping \mbox{($b=0.02$, $p=.64$)}. However, as lower-distance pairings were excluded, the effect grew in magnitude and became statistically significant: restricting to the top 50\% most distant pairings yielded a significant effect \mbox{($b=0.12$, $p=.029$)}, and restricting to the top 10\% produced a substantially larger effect \mbox{($b=0.24$, $p=.024$)}. Together, these results suggest that the benefit of cross-domain mapping scales with semantic distance for both humans and LLMs, but the distance required before a benefit appears differs considerably across humans, LLMs on average, and the best LLMs. One possibility is that LLMs already access relatively remote associations under the user-need prompt, leaving less additional benefit for moderately distant supplied sources; we revisit this interpretation in the Discussion. 

\subsection{Originality Comes With a Price for Humans, but Less So for LLMs}

The preceding analyses establish that LLMs produce more original ideas than humans on average, but human ideas at the upper tail---particularly under cross-domain mapping---can match or even exceed LLM-generated ideas on originality. Here, we note a trade-off affecting humans but not LLMs: when humans reach the frontier of originality, their ideas come at a cost to other dimensions in ways that LLM ideas do not.

\begin{figure}[t!]
\centering
\includegraphics[width=0.95\textwidth]{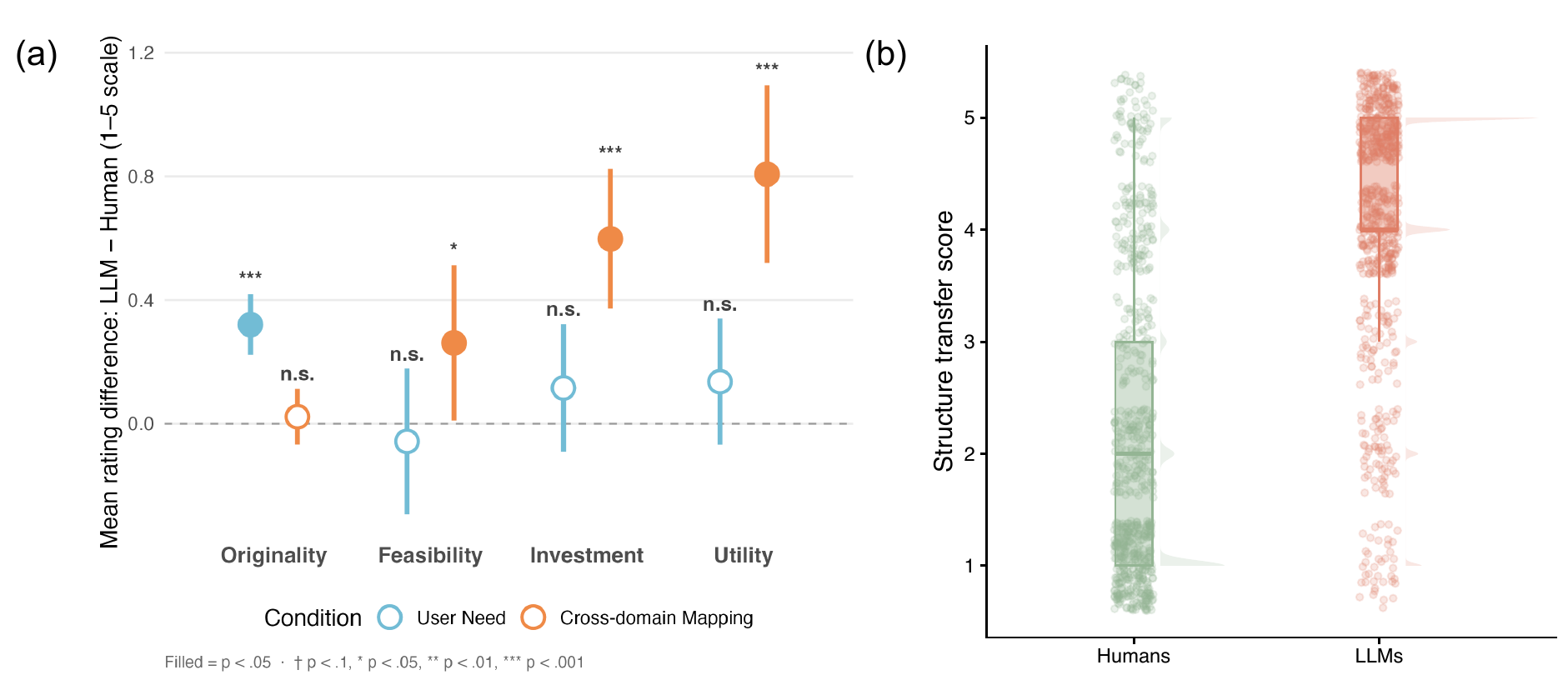}
\caption{Originality--utility trade-off and surface-vs-structure transfer. (a) Mean rating differences (LLM $-$ human, 1--5 scale) on the top 10\% most original ideas in each pool, by condition. Points are coefficients from linear mixed-effects models (positive = LLM rated higher); filled = $p < .05$; error bars = 95\% CIs. $^{*}p < .05$, $^{**}p < .01$, $^{***}p < .001$, $^{\dagger}p < .10$. (b) Manually labeled structure-transfer scores for all 1400 cross-domain ideas (1 = pure surface transfer, 5 = pure structural transfer).}
\label{fig:structural}
\end{figure}

We identified the 10\% most original ideas generated by humans and LLMs respectively. We then compared their ratings across all four dimensions using linear mixed-effects models with random intercepts for raters and ideas. The results are shown in Figure~\ref{fig:structural}a, where positive coefficients for a dimension indicate that LLM ideas were rated higher along that dimension than human ideas.

The results reveal an interesting asymmetry between the two prompting conditions. In the \textit{user-need} condition, the top 10\% of human ideas were rated as less original than their LLM counterparts ($b = 0.32$, $SE = 0.05$, $p < .001$), yet matched LLM ideas on utility ($b = 0.14$, $p = .20$), feasibility ($b = -0.06$, $p = .63$), and investment-worthiness ($b = 0.12$, $p = .28$). In other words, when humans focus on addressing user needs, their most original ideas were less novel than the LLMs' but just as useful, feasible, and worthy of investment.

Under \textit{cross-domain mapping}, this pattern flipped. The top 10\% of human ideas now matched LLMs on originality ($b = 0.02$, $p = .63$), but were rated as substantially less feasible ($b = 0.26$, $p = .045$), less worthy of investment ($b = 0.60$, $SE = 0.12$, $p < .001$), and most prominently, less useful ($b = 0.81$, $SE = 0.15$, $p < .001$). Some of these highly original but impractical human ideas include a desk inspired by a \textit{pickle} (``This desk is green, gives off a vinegar smell, and releases juice when you squeeze it,'' originality = 4.78/5, utility = 1.22/5) or a TV inspired by a \textit{sandwich} (``This TV has a bread-shaped frame with layered sections that look like sandwich fillings...''; originality = 4.75/5, utility = 1.13/5).

The most original human ideas come at a sharp cost to utility, while equally original LLM ideas do not. Why does cross-domain inspiration lead humans, to trade away utility? Humans and LLMs may be drawing on source domains in fundamentally different ways and transferring different kinds of information from source to target. We investigate this possibility next.

\subsection{Humans Transfer Surface Features; LLMs Transfer Structures and Functions}

A long-standing distinction in the analogy literature holds that mappings vary in depth: one can import a source's surface appearance, or one can import its underlying structure \citep{gentner1986systematicity}. This distinction offers a natural explanation for why cross-domain inspiration might lead humans, but not LLMs, to sacrifice utility for originality: the two systems may favor different kinds of transfer.

As a first, indirect test, we looked for a lexical signature of surface-based mapping using the Lancaster Sensorimotor Norms \citep{lynott_lancaster_2020}, which quantify how strongly each English word evokes perceptual experience (e.g., visual, auditory, haptic) and bodily action. For each content word in an idea, we retrieved its maximum sensorimotor strength across the eleven perceptual and action dimensions in the norms, and averaged these values across all matched content words to obtain an idea-level sensorimotor score. In the cross-domain mapping condition, human ideas drew more on words that evoke sensory and motor experience than LLM ideas ($M = 3.11$ vs.\ $3.07$, $p < .001$). Crucially, this is not a general human preference for perceptual vocabulary, because in the user-need condition, where no source domain is provided, the direction reverses and human ideas rely on \emph{less} sensory and action-laden language than LLM ideas ($M = 3.02$ vs.\ $3.08$, $p < .001$; see also Supplemental Material, Figure~S7).

To directly measure the extent to which humans and LLMs transfer structure in cross-domain mappings, the first author manually coded all 1400 cross-domain ideas on a 1--5 scale, from 1 indicating pure surface transfer (importing a visual or physical attribute without its function) to 5 indicating pure structural transfer (importing a causal mechanism or organizational principle without surface resemblance). LLM ideas clustered near the structural end of the scale, while human ideas were concentrated at the surface end of the scale, with a minority reaching higher structural scores, suggesting that deep analogical mapping is possible but effortful for our human participants (Figure~\ref{fig:structural}b). For example, inspired by pickles, many human participants proposed a car with ``pickle-green paint with grooves and bumps,'' or inspired by cacti, a backpack with thorny textures to deter theft. In contrast, LLM cross-domain mappings more often align and project mechanisms or functions across domains: the same example of a pickle-inspired car leads to a self-healing, anti-corrosion mechanism (micro-brine capsules that seal scratches), while the cactus becomes a water-harvesting system (micro-spine condensation fins channeling moisture into a drinkable reservoir). It's unlikely that LLMs don't know pickles are typically green and dimpled while cacti are spiky, but they differ from humans in what is treated as generative during analogical transfer.

This surface-vs-structure distinction offers an explanation for why human ideas in the cross-domain mapping condition can reach high levels of originality yet tend to be less useful. We examined how the structure-transfer score related to each rating dimension within human ideas using linear mixed-effects models with random intercepts for raters and ideas. We found the structure-transfer score was weakly associated with originality ($b = 0.06$, $p = .026$), but was more reliably associated with utility ($b = 0.21$, $p < .001$) (Full results across all four rating dimensions for both human and LLM ideas are reported in Supplemental Material, Figure~S8). 


\section{Discussion}
\label{sec:discussion}

Creativity is often associated with sudden and serendipitous cross-domain inspirations. The central question motivating this work was whether such cross-domain mappings can be induced at scale, and whether cross-domain mappings affect creative insights differently for humans and LLMs. We found that (1) LLMs generated ideas that were perceived as more original on average, regardless of whether cross-domain mappings were explicitly prompted; (2) on average, cross-domain mapping prompts increased the perceived originality of human but not LLM ideas; (3) across both systems, greater semantic distance between target and source domains predicted higher perceived originality, and once that distance was sufficiently large, the most capable LLMs also benefited from cross-domain mapping; and (4) humans and LLMs used source domains in qualitatively different ways, with consequences for the utility of the resulting ideas.

\subsection{Why the Benefits of Cross-Domain Mapping Differ Across Humans and LLMs}

Consistent with previous work showing that responses farther from their cues in semantic space tend to be judged as more original, we found that more distant target--source pairings yielded more original ideas for both humans and LLMs. The pairings that were productive for humans also tended to be productive for LLMs. What differed was the distance at which supplying a source outperformed the user-need baseline. For the human participants in our study, almost any source helped. For tested LLMs as a group, cross-domain mapping performed worse than the user-need prompt for close pairings and only caught up for more distant ones. For the two highest-rated models, \texttt{o3} and \texttt{Claude-Sonnet-4}, a benefit emerged among the more distant half of the pairings.

One possible explanation is that an external source helps only when it pushes a system beyond the region of semantic space it would readily explore. Without a supplied source, human participants tended to remain close to the target, so even a moderately remote source could push them beyond incremental improvements to existing functions. An LLM may reach a broader region of semantic space without any external source because it has been exposed to far more linguistic information than any individual person. Even at the high end of estimates of human linguistic exposure, a person encounters roughly 1.7 billion words by age 60 \citep{brysbaert2016many,wingfield2022understanding}. In contrast, the three open-weight models in our study were pretrained on 15--18 trillion tokens \citep{grattafiori2024llama,qwen2024qwen25,deepseekai2024v3}, equivalent to roughly 6,600--7,900 human lifetimes of linguistic exposure. The comparison is necessarily approximate, and people also learn through perception, action, and social experience, but it illustrates the extraordinary scale of semantic information available during pretraining. Concepts that are remote in an individual's experience may therefore be easier for an LLM to reach without external guidance, leaving less for a supplied source to add.

More semantic knowledge alone, however, cannot explain why distant sources especially benefited \texttt{o3} and \texttt{Claude-Sonnet-4}. The pretraining corpora of these closed-weight models are undisclosed, but simply assuming that they performed better because they were trained on more data would actually predict the opposite pattern---If being pretrained on more data makes a wider range of concepts accessible to a model, then supplying an external source should provide less additional benefit, not more. The advantage of these models may therefore lie not only in what knowledge they have available, but also in how effectively they use a remote source once it is supplied. This distinction is similar to a distinction made in executive theories of human creativity, which separate the structure of semantic knowledge from the processes used to search and use that knowledge. On this view, \texttt{o3} and \texttt{Claude-Sonnet-4} may be especially effective at constructing useful mappings from remote concepts. Consistent with this possibility, differences among models were larger under cross-domain mapping than under the user-need prompt. Determining whether this advantage reflects mapping ability, semantic knowledge, or some combination of the two will require methods that probe the generative process rather than inferring it from outputs alone. Under either account, we think the human--LLM asymmetry is better interpreted as a difference in sensitivity to cross-domain intervention than as evidence that cross-domain mappings are less generative for LLMs in principle.

\subsection{Originality, Feasibility, and Utility}

Across ideas, originality showed a strong negative relationship with feasibility and a weaker but still negative relationship with utility, especially among human-generated ideas. At face value, this might suggest that pushing ideas toward high originality necessarily produces impractical ideas with unclear utility. Some of the most original human ideas we collected illustrate the concern: a desk that smells of vinegar and oozes juice when squeezed or a TV with a sandwich-shaped frame layered to mimic fillings. These ideas are vivid and unexpected, but it is hard to see what problem they solve.

But this does not mean every impractical-looking idea should be dismissed. Judgments of feasibility and utility are necessarily made relative to a currently imaginable problem/solution space \citep{shtulman2023learning}. Ideas that sit outside that space may look both impossible and pointless, not necessarily because they violate principles of physics or markets, but because the mechanisms by which they could be realized are not yet representable. Many historically transformative ideas initially suffered from this problem. For example, early skepticism about AI-generated arts often rested on the assumption that a creative machine would entail a human-like robot arms manipulating brushes \citep{mccorduck1991aaron}, rather than statistical models operating over latent representations---an implementation path that was not merely unlikely, but effectively \textit{unthinkable} at the time. Once a workable path appears, judgments of feasibility can shift rapidly; and until adoption, utility is necessarily speculative.

\subsection{Embodiment, Language, and the Structuralness of Analogy}

We observed a clear qualitative difference between humans and LLMs in what they transferred from a source given a cross-domain mapping: humans tended to transfer surface features, while LLMs tended to transfer relational structure and function. This difference speaks to the broader question motivating this work: which principles of creativity generalize across intelligent systems, and which are tied to the particular constraints of being human?

A simple explanation is that human participants put limited effort into the task, while LLMs, by design, give their best response to the questions they are posed. We cannot objectively measure the effort put in by participants, but we can instruct LLMs to try less. In a preliminary investigation, when we instructed several LLMs to respond as a lazy human who doesn't think too much, their outputs grew shorter but remained predominantly structural. When we instead instructed them to respond as an artist attentive to perceptual features, outputs shifted toward surface mappings, and pickle-green paint cars are proposed. The gap, then, may be less about effort than about which features each system is steered to attend to.

Another explanation lies in differences in domain knowledge. As reviewed earlier, expertise can help people look beyond surface similarity and recognize deeper structural correspondences. But expertise is usually domain-specific. Our inspiration sources spanned animals, plants, foods, technologies, natural systems, and social organizations, so even a participant with deep knowledge in one area would likely be a novice in many others. LLMs, by contrast, have broad cross-domain knowledge, including the structural and functional properties of random source concepts. Therefore, what would require bringing together people with different areas of expertise can, in an LLM, be available within a single system.

A deeper explanation is that the surface bias documented in decades of analogy research \citep{gentner1993roles,holyoak1987surface,ross1987this} reflects the conditions under which human cognition operates. For humans, many concepts are anchored more deeply in embodied perception than in abstract description: a pickle is something we have seen, held, and tasted far more often than we have read about its fermentation chemistry. Perceptual features are immediate and cheap to retrieve, while the relational structure that facilitates useful inference requires effortful abstraction and structural alignment. LLMs, by contrast, never experience a pickle perceptually. They encounter the concept primarily through text in which relational and functional properties (\emph{how} they are produced, \emph{why} they preserve well, etc.) are richly described, while perceptual features may be comparatively underspecified. This account is consistent with evidence that people who lack direct perceptual access to a given modality shift toward relational and functional content. Congenitally blind adults, for instance, list reliably fewer perceptual features of concrete objects (e.g., that a pencil is cylindrical) and more contextual, taxonomic, and functional ones (e.g., that a pencil is used with paper) than sighted adults \citep{lenci2013blind,kim2019knowledge,mamus2025gestural}.

\subsection{Limitations and Future Directions}

First, our design captures one-shot idea generation, whereas real-world creativity is inherently iterative: ideas are criticized, combined, revised, and gradually made workable. Future work should therefore study iterative workflows which could instantiate explicit mechanisms that humans naturally use in creative practice---e.g., reflection and adversarial critique that could be implemented either within models (self-critique and refinement loops), in multi-agent settings, or in human--AI collaborations. 

Second, our originality, feasibility, and utility ratings come from lay judges, which is appropriate for consumer products but leaves unanswered whether the same evaluations would generalize to other rater populations or domain experts, or predict what consumers would actually buy. LLM raters have been used for such judgments \citep{patterson2025cap} and are far cheaper to scale, but they are known to favor machine-generated text \citep{panickssery2024llm,wataoka2024self}, a bias that would directly confound the human--LLM comparison which is the focus of this study. A human-in-the-loop approach, in which LLM ratings are calibrated against and audited by human judges, is one direction future work could take.

Third, our findings may not generalize to other creativity tasks or to other populations. Our task may favor systems with broad knowledge across many domains, where LLMs have a clear advantage, but the same pattern may not hold for tasks that require less domain knowledge but rely more on taste, emotion, or lived experience, such as poetry, humor, or visual art. In those tasks, perceptual or embodied associations may contribute to emotional resonance or aesthetics in ways that matter less for product design. The human--LLM differences we observed may also vary across populations with different levels of expertise, education, or experience with the source domains.

Finally, we observe what each system produced, not the candidate mappings it considered and discarded. Our findings therefore characterize differences in observed outputs, including how cross-domain mapping affected originality and what kinds of source properties humans and LLMs ultimately transferred, rather than the underlying generative processes that produced those differences. Probing that process, e.g., through think-aloud protocols for humans, and reasoning traces, internal representations, or the distribution of sampled completions for LLMs, is a natural next step we leave to future work.

\section*{Author Note}

Marina Dubova and Henry Conklin contributed equally to this work.

Q.E.L. and T.L.G. conceived the project and designed the methodology. Q.E.L. performed data collection, analysis, visualization, and wrote the initial draft. M.D. contributed to conceptualization and analytical approach through extensive discussion throughout the project. H.C. developed the semantic distance metric and drafted the relevant description. T.L.G. provided project administration and supervision. Q.E.L., M.D., H.C., T.H., and T.L.G. reviewed and edited the manuscript.

The authors declare no competing interests. We thank Toyota Motor North America, Inc.\ for their support of this research. We also thank Pavel Chvykov, Matthew Henderson, and the Diverse Intelligence Summer Institute for many thoughtful discussions that inspired this work.

Correspondence concerning this article should be addressed to Qiawen Ella Liu, Princeton University. E-mail: ella.qiawen.liu@princeton.edu

\section*{Data Availability}

This research is approved by the Princeton University Institutional Review Board (Protocol No.~17087). Data, materials, and preregistrations are available through the Open Science Framework at \href{https://osf.io/xp4ed/overview?view_only=15510d5b15ba488fb6bcd0354ad77ba4}{OSF (https://osf.io/xp4ed/)}.

\begingroup
\renewcommand{\bibliographytypesize}{\small\linespread{1.0}\selectfont}
\setlength{\bibitemsep}{3pt}
\bibliographystyle{apacite}
\bibliography{references}
\endgroup

\end{document}